\documentclass[conference]{IEEEtran}
\usepackage[pdftex]{graphicx}
\usepackage{graphicx}
\usepackage{amsmath}
\usepackage{fixltx2e}
\usepackage{multirow}
\usepackage{textcomp}

\ifCLASSINFOpdf
\else
\fi
\usepackage{array}
\usepackage{mdwmath}
\usepackage{mdwtab}
\usepackage{eqparbox}
\usepackage[tight,footnotesize]{subfigure}
\begin{document}

\title{Thinning Algorithm Using Hypergraph Based Morphological Operators  }

\author{\IEEEauthorblockN{Prakash R P}
\IEEEauthorblockA{PG Student\\
 Computer Science and Engineering\\
College Of Engineering\\Karunagappally, India\\prakasharumanoor@gmail.com}
%\hspace{8cm}
\and
\IEEEauthorblockN{Keerthana S Prakash}
\IEEEauthorblockA{PG Student\\ Computer Science and Engineering\\
	College Of Engineering\\Karunagappally, India\\keerthanatheertha@gmail.com}
\and
\IEEEauthorblockN{Binu V P}
\IEEEauthorblockA{Associate Professor
	\\Computer Science and Engineering\\
	College Of Engineering\\Karunagappally, India\\
binuvp@gmail.com}
}


%\author{\IEEEauthorblockN{Keerthana S Prakash\IEEEauthorrefmark{1},
%Prakash R P\IEEEauthorrefmark{2},
%Binu V P\IEEEauthorrefmark{3} and 
%Bino Sebastian V\IEEEauthorrefmark{4}}
%
%\IEEEauthorblockA{\IEEEauthorrefmark{1}PG Student, Computer Science and Engineering, College Of Engineering, Karunagappally\\ keerthanatheertha@gmail.com}
%\IEEEauthorblockA{\IEEEauthorrefmark{2}PG Student, Computer Science and Engineering, College Of Engineering, Karunagappally\\ prakasharumanoor@gmail.com}
%\IEEEauthorblockA{\IEEEauthorrefmark{3}Associate Professor, Computer Science and Engineering, College Of Engineering, Karunagappally\\
%	binuvp@gmail.com}
%\IEEEauthorblockA{\IEEEauthorrefmark{4}Research Scholar, Department of Computer Applications, Cochin University of Science and Technology\\
%		binosebastianv@gmail.com}
%
%}


\maketitle

\begin{abstract}

The object recognition is a complex problem in the image processing. Mathematical morphology is Shape oriented operations, that simplify image data, preserving their essential shape characteristics and eliminating irrelevancies. This paper briefly describes morphological operators using hypergraph and its applications for thinning algorithms. The morphological operators using hypergraph method is used to preventing errors and irregularities in skeleton, and is an important step recognizing line objects. The morphological operators using hypergraph such as dilation, erosion, opening, closing is a novel approach in image processing and it act as a filter remove the noise and errors in the images.

\end{abstract}
%\begin{keywords}
	
%	{
		
%		Mathematical morphology, hypergraph, thinning,
%		skeleton
		
%	}
	
%\end{keywords}

 \textbf{Index Terms}-Mathematical morphology, graphs, hyper graph, alternative sequential filter



% page as needed:and Montgomery Scot
% \fi
\IEEEpeerreviewmaketitle

\section{Introduction}
Recognition of line objects is complex problem that can be solved in many ways. It consists of more phases depending on approach. Each phase affects next so it is important to obtain good results after the first ones. Pre-processing is the first step in all methods. It modifies input raster to enhanced important information and wipe out those that can cause future problems (like noise). In the next step we usually use some type of thinning to create skeleton. So the image become  reducing all lines to single pixel thickness. There are many approaches on how to create skeleton with different results. Accuracy of results heavily depends on input quality and characteristics. This paper focuses on the morphological operations using hypergraph and thinning process can be applied, which is essential for many image processing tasks including line objects recognition.\par
Mathematical morphology \cite{3},  appeared in 1960s, mainly based on set-theoretic, and geometric principles. Mthematical morphology is developed by Matheron and Serra, mainly for binary images. There has been not more work done on the hypergraphs. This technique is used to extract characteristic features of the images which are useful for specific applications. Morphological operations are developed on graphs and hypergraph. Graph \cite{6}, is collection of vertices and edges. The binary relation between the vertices are called edges. Thus binary relations between the objects in the image can be represented by using graph \cite{15}. \par
Hypergraph theory, which is developed by C.Berge  in 1960. Hypergraph is a generalization of graphs. In hypergraph more two nodes can be connected at a time. But in the case of graph at a time two nodes can be connected. So there is more connection between nodes hypergraph gives more information than a graph strcture. Based on the connectivity between the nodes in a hypergraph, there are different type of hypergraph. They are 2-uniform hypergraph \cite{14}, connect two nodes at a time.Basically it is a graph structure. In a 3-uniform hypergraph is a collection of three nodes and so on. The rank of a hyperedge is the number of vertices incident with that edge.The representation of hypergraph can be shown in Figure 1. Here M1, M2, M3, M4 are the edge set and each edge set contain one ore more than one vertex. For example edge M2 contain vertex v2 and v3 and edge M4 contain vertices v3 and v5.   \par
In this paper, we explore how morphological operators can be applied to image pre processing before thinning so that better result can be obtained.  
 \begin{figure}[h]
 	\centering
 	\includegraphics[width=0.50\textwidth]{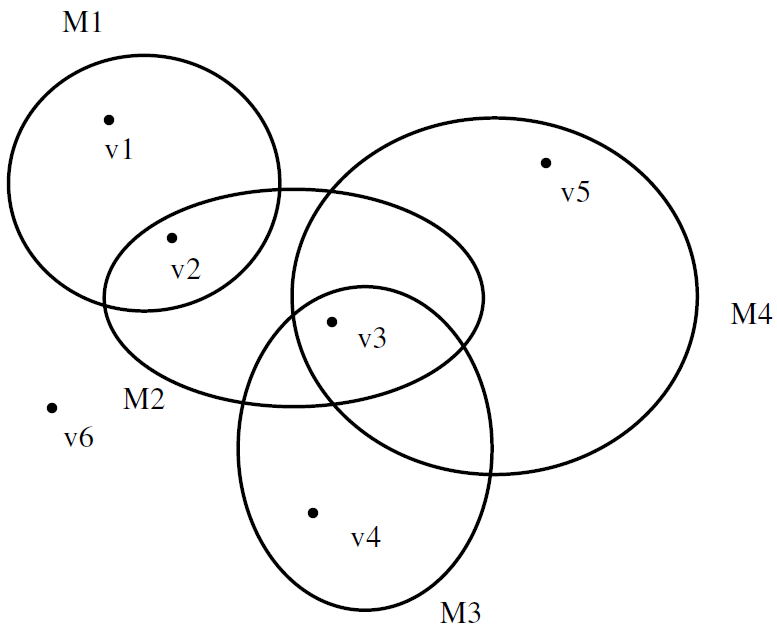}
 	\caption{Example of hypergraph}
 	\label{fig:}
 \end{figure}\par
  \section{Basic theoretical concepts}
  \subsection{Mathematical Morphology}
  Mathematical Morphology is a theory which provides a number of useful tools for image analysis.based on set theory\cite{3}.Mathematical morphology is an approach to image analysis which is based on the set theory.  It use the set-theoretical operations like union and intersection. In order to apply it to  grey-level image, it is necessary to generalize set-theoretical notions. All the basic morphological operators are defined by using this framework.	Morphological techniques uses image with a small shape known as a structuring element. It can be of any shape or size.
  
  \subsection{Hypergraph}
  Hypergraph \cite{5} can be represented as a pair    $H=(H^{\bullet},H^{\times})$.Here $H^{\bullet}$ represent the vertices and $H^{\times}$ subset  $H^{\bullet}$
  called hyperedges. $H^{\times}$ can be represented as $(e^{i})$ where I is the finite set of indices. of H if $x\notin \underset{i\in I}{\bigcup} v(e_{i})$ Vertex set forming the hyperedge e is called v(e).Vertex x of $X^{\bullet}$ is said to be isolated vertex. \par
   A subhypergraph $H_{a}$  induced by a subset A of X is defined as $H_{A}=(A,{e_{i} \cap A)\rvert e_{i}\cap A\neq 0})$. The partial hypergraph is a hypergraph with some edges removed. \par
  
    A subset $A \subseteq X $, $H\times A=(A,{e_{i}}|i\in I_{e},e_{i}\subseteq A)$.The dual $H^{\ast}$  of H is a hypergraph whose vertices and edges are interchanged, so that the vertices are given by ${e_{i}} $ and whose edges are given by ${X_{m}} $ where $X_{m}={e_{i}|x_{m}\in e_{i}}$.

\section{Morphological operations on hypergraph}
Complex relations in the images can be represented by using hypergraph. Bino et al. [10], [11] defined morphological operators on hypergraph. Hypergraph consist of sets of points as well as sets of hyperedges. It is convenient to consider basic operators to go from one kind of sets to the other one. \\
\\
\textbf{Property 1}. [10] For any $ X^{\bullet} \subseteq H^{\bullet} $ and any $ X^{\times} \subseteq H^{\times} $, where $ X^{\times}=(e_{j} ) $, $j \in J$
such that $ J\subseteq I $\\
\begin{enumerate}
	\item  $\delta^{\bullet} \colon H^{\times} \rightarrow H^{\bullet}$ is such that $ \delta^{\bullet}(X^{\times})=\underset{j \in J}{\cup} v(e_{j})$
	\item{ $\epsilon^{\times}\colon H^{\bullet} \rightarrow H^{\times}$  is such that \\$ \epsilon^{\times}(X^{\bullet}) =\{{e_{i},i \in I|v(e_{i})\subseteq X^{\bullet}}$}\}
	\item {$ \epsilon^{\bullet}\colon H^{\times} \rightarrow H^{\bullet} $ is such that $ \epsilon^{\bullet}(X^{\times})=\underset{j\notin J}{\cap} 
		\overline{v(e_{j})} $} 
	\item{ $\delta^{\times}\colon H^{\bullet} \rightarrow H^{\times}$  is such that\\ $ \delta^{\times}(X^{\bullet}) =\{{e_{i},i \in I|v(e_{i}) \cap X^{\bullet}\neq\emptyset}$}\}.
	\\
	
\end{enumerate}
\textbf{Property 2}. [10] (dilation, erosion, adjunction, duality)\\
\begin{enumerate}
	\item {Operators $\epsilon^{\times}$ and $\delta^{\times}$ (resp. $\epsilon^{\bullet}$ and $\delta^{\bullet})$ are dual of each other}.
	\item Both $(\epsilon^{\times}$, $\delta^{\bullet})$ and $(\epsilon^{\bullet},\delta^{\times})$ are adjunctions.
	\item Operators $\epsilon^{\bullet}$ and $\epsilon^{\times}$ are erosions.
	\item Operators $\delta^{\bullet}$ and $\delta^{\times}$ are dilations.
	\\
\end{enumerate}
\textbf{Definition 1. [10] (vertex dilation, vertex erosion)}
$\delta$ and $\epsilon$ defined that act on $H^{\bullet}$ by $\delta=\delta^{\bullet}\circ\delta^{\times}$ and $\epsilon=\epsilon^{\bullet}\circ\epsilon^{\times}$.\\
\\
\textbf{Property 3.} [10] For any $X^{\bullet} \subseteq H^{\bullet}$\\
\begin{enumerate}
	\item $\delta(X^{\bullet})=\{x\in H^{\bullet}| \exists e_{i}, i\in I$ such that \\$x\in v(e_{i})$ and $v(e_{i})\bigcap X^{\bullet}\neq\emptyset\}.$
	\item $\epsilon(X^{\bullet})=\{x\in H^{\bullet}| \exists e_{i}, i\in I$ such that \\$x\in v(e_{i})$ and $v(e_{i})\subseteq X^{\bullet}\}.$
	\\
\end{enumerate}
\textbf{Definition 2. [10] (hyper-edge dilation, hyper-edge erosion)} $\Delta$ and $\varepsilon$ defined that act on $H^{\times}$ by $\Delta=\delta^{\times}\circ\delta^{\bullet}$ and $\varepsilon=\epsilon^{\times}\circ\epsilon^{\bullet.}$
\\
\\
\textbf{Property 4.} [10] For any $X^{\times}\subseteq H^{\times}, X^{\times}=(e_{j})_{j\in J}$
\begin{enumerate}
	\item $\Delta(X^{\times})$=$\{e_{i},i\in I|\exists e_{j},j\in J$ such that \\
	$ v(e_{i})\cap v(e_{j}\neq\emptyset)\}.$
	
	\item $\varepsilon(X^{\times})=\{e_{j},j\in J|v(e_{j})\cap v(e_{i})\neq\emptyset,\forall i\in I, J \}.$
	\\
\end{enumerate}
\textbf{Definition 3. [10] (hypergraph dilation, hypergraph erosion)} The operators $[\delta,\Delta]$ and $[\epsilon,\varepsilon]$ defined by respectively $[\delta,\Delta](X)=(\delta(X^{\bullet}),\Delta(X^{\times}))$ and $[\epsilon,\varepsilon](X)=(\epsilon(X^{\bullet}),\varepsilon(X^{\times}) ),$ for any $x\in H.$
\\
\\
\textbf{Definition 4. [11] (opening, closing)}
\begin{enumerate}
	\item $\gamma_{1}$ and $\varPhi_{1}$ defined, that act on $H^{\bullet}$, by $\gamma_{1}=\delta \circ \epsilon$ and $\varPhi_{1}=\epsilon \circ \delta.$
	
	\item $\Gamma_{1}$ and $\Phi_{1}$ defined, that act on $H^{\times}$, by $\Gamma_{1}=\Delta\circ\varepsilon$ and $\Phi_{1}=\varepsilon \circ
	\Delta.$
	\item $[\gamma,\Gamma]_{1}$ and $[\varPhi,\Phi]_{1}$ defined, that act on H by respectively $[\gamma,\Gamma]_{1}(X)=(\gamma_{1}(X^{\bullet}),\Gamma_{1}(X^{\times}))
	$and $[\varPhi,\Phi]_{1}(X)=(\varPhi_{1}(X^{\bullet}),\Phi_{1}(X^{\times}))$ for any $X\in H.$
	
\end{enumerate}
\textbf{Definition 5. [11] (half-opening, half-closing)}\\
\begin{enumerate}
	\item $\gamma_{1/2}$ 
	and $\varPhi_{1/2}$ defined, that act on $H^{\bullet}$, by $\gamma_{1/2}=\delta^{\bullet} \circ \epsilon^{\times}$ and $\varPhi_{1/2}=\epsilon^{\bullet} \circ \delta^{\times}.$
	
	\item $\Gamma_{1/2}$ and $\Phi_{1/2}$ defined, that act on $H^{\times}$, by 
	$\Gamma_{1/2}=\delta^{\times}\circ\epsilon^{\bullet}$ and $\Phi_{1/2}=\epsilon^{\times} \circ
	\delta^{\bullet}.$
	\\
\end{enumerate}
\textbf{Property 5. [11] (hypergraph opening, hypergraph closing)}\\
\begin{enumerate}
	\item The operators $\gamma_{1/2}$ and $\gamma_{1}$ (resp. $\Gamma_{1/2}$ and $\Gamma_{1})$ are opening on $H^{\bullet}$ (resp. $H^{\times}$) and $\varPhi_{1/2}$ and $\varPhi_{1}$ (resp. $\Phi_{1/2}$ and $\Phi_{1})$ are closing on $H^{\bullet}$.
	
	\item The family H is closed under $[\gamma,\Gamma]_{1/2},[\varPhi,\Phi]_{1/2},$ $[\gamma,\Gamma]_{1},[\varPhi,\Phi]_{1}.$
	\item $[\gamma,\Gamma]_{1/2}$ and $[\gamma,\Gamma]_{1}$ are opening on H and $[\varPhi,\Phi]_{1/2}$ and $[\varPhi,\Phi]_{1}$ are closing on H.\\
	
\end{enumerate}

\section{MORPHOLOGICAL APPLICATIONS}
Recognition of line objects is a specific process, which deals with two main problems:
\begin{enumerate}
\item Connectivity preserving. If two objects are connected in original image, they
must be connected also after recognition and the connectivity should be preserved.

\item Shape preserving. Although it is not important to preserve exact shape or proportions, characteristics of shape should be preserved.

 \end{enumerate} 
Probably the best way how to deal with these two requirements is to use thinning to create skeleton. The Zhang-Suen thinning algorithm \cite{2} is applied here to show how input errors and uncertainties influence the skeleton.\par
As shown in Figure 2, inaccurate input can cause not only bad shape recognition but also what is worse; it can change connectivity of objects. Input picture consists of one object that has shape “O”. The desired skeleton consists of one line object that preserves this shape and also preserves connectivity (it is not connected with other
object). On the other hand, skeleton which was acquired without any pre-processing, like any morphological operation like dilation and erosion consists of more objects that hardly describe original shape and it does not preserve connectivity of its origin at all.
\par

Figure 3 demonstrate the possibilities of using morphological operators dilation(D) and erosion(E) using hypergraph in pre-processing phase to eliminate most common errors like noise, line fuzz, holes and separations.

\begin{figure}[h]
	\centering
	\includegraphics[width=0.50\textwidth]{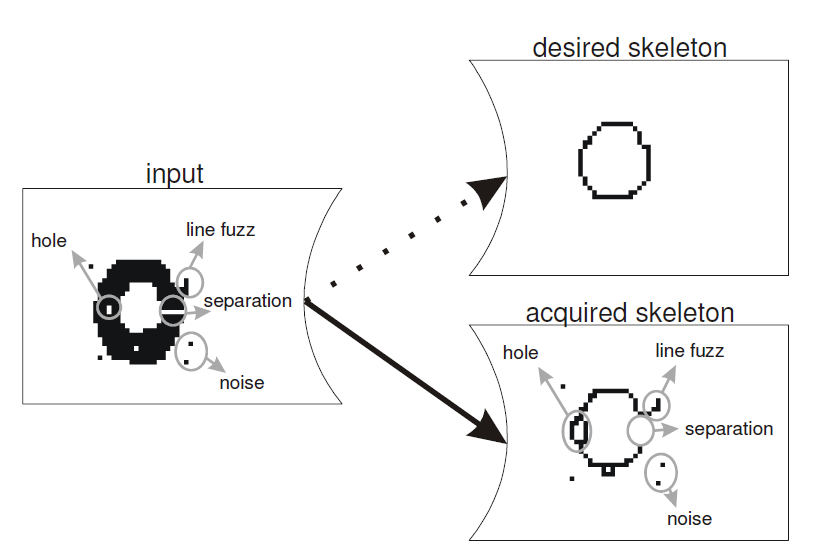}
	\caption{Input, desired skeleton and acquired skeleton (without any pre-processing)}
	\label{fig:}
\end{figure}
\begin{figure}[h]
	\centering
	\includegraphics[width=0.50\textwidth]{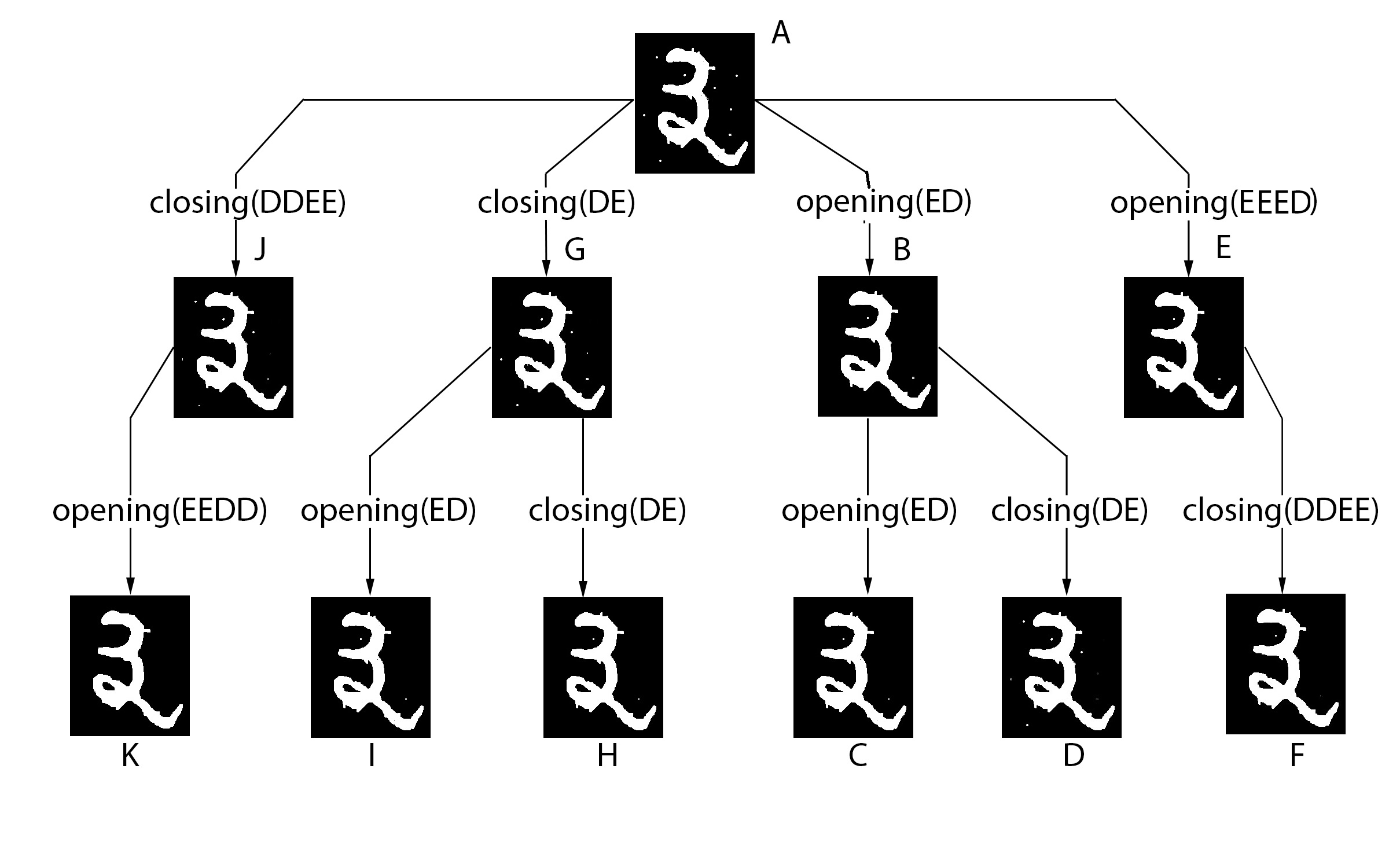}
	\caption{Some possible results of binary morphology}
	\label{fig:}
\end{figure}

Picture A is binary input from which it is possible to get pictures J, G, B and E by using different number and order of dilation and erosion operations using hypergraph. While only one  hypergraph dilation and erosion was used in pictures G and B. Pictures J and E show possibilities of using higher number of the same morphological operations using hypergraph one after another.

Opening operation using hypergraph is good for noise reduction and line fuzz removal. Picture B
shows that all noise and small line fuzz was removed. Its effect depends on number of iterations ie, number of dilation and erosion as it is shown in picture E, where two iterations were used. If too much iteration is used there is a danger of erasing real parts of the object (line fuzz) or whole objects at all (noise). Similar situation is shown in pictures G and J, where  hypergraph closing  is used. This operation causes holes reduction. Also there is a threat of using too much iteration as it can fill objects that can lead to loss of their line characteristic Morphological operations like dilation, erosion using hypergraph in pre-processing phase to eliminate most common errors like noise, line fuzz, holes and separations.

\section{ EXPERIMENTAL RESULTS }

The effect of hypergraph based morphological operators in pre processing is shown with the binary noise image. The result of thinning with or without hypergraph based morphological operators is studied. Figure 4, is the original noisy image and the result of thinning image without morphological operations are shown in figure 5. But the resultant image which contain some amount of noise. Consider the figure 7, which is the resultant image of thinning using morphological operators, is better than the previous result. So the effect of morphological operation which leads to eliminate most common errors like noise, line fuzz, holes and separations. On the other hand,skeleton, which was acquired without any preprocessing, consists of more objects that hardly describe original shape and it does not preserve connectivity of its original at all.

\begin{figure}[!htb]\centering
	\begin{minipage}{0.20\textwidth}
		\includegraphics[width=\linewidth]{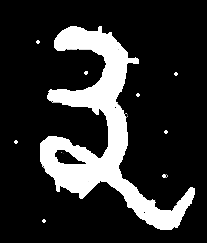}
		\caption{Original Image}
		\label{fig1}
	\end{minipage}
	\hspace{0.5cm}
	\begin {minipage}{0.20\textwidth}
	\includegraphics[width=\linewidth]{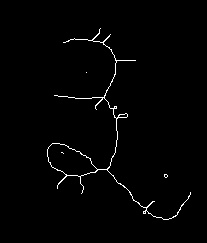}
	
	\caption{thinning without morphological operation}
	%	\label{fig2}

\end{minipage}
%	\caption{Original test image and its noisy version}
\end{figure}

\begin{figure}[!htb]\centering
	\begin{minipage}{0.20\textwidth}
		\includegraphics[width=\linewidth]{prak}
		\caption{original image}
		\label{fig1}
	\end{minipage}
	\hspace{0.5cm}
	\begin {minipage}{0.20\textwidth}
	\includegraphics[width=\linewidth]{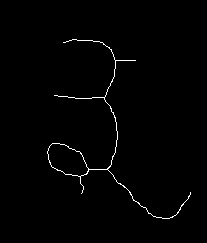}
	\caption{Result of thinning with morphological operation}
	\label{fig2}
\end{minipage}
\end{figure} 

\section{Conclusion}
We studied the different morphological operations on the hypergraph. Thinning process which can be performed on the images using morphological operators using hypergraph and which gives noise free skeletonized image as output. The thinning using morphological operators shows better results than compared with thinning without using morphological operators.The morphological operations have many application in the satellite imaging, medical imaging technique etc. The technique can be used as the effective preprocessing step in the recognition of handwritten characters, number plate detection.





%\bibliographystyle{IEEEtran}

\newpage
\begin{thebibliography}{1}








\bibitem{1}TARÁBEK, Peter, "Morphology image pre-processing for thinning algorithms", \emph{ Journal of Information, Control and Management Systems}, vol. 5, no. 1, 2007

\bibitem{2}Zhang, T. Y., and Ching Y. Suen, "A fast parallel algorithm for thinning digital pattern," \emph{Communications of the ACM}, vol. 27, no. 3,  pp. 236-239, 1984

\bibitem{3} Heijmans, Henk JAM and Christian Ronse, "The algebraic basis of mathematical morphology I. Dilations and erosions, \emph{Computer Vision, Graphics, and Image Processing}, vol. 50, no. 3, pp. 245-295, 1990.


\bibitem{4}Gonzalez, Rafael C., and Richard E. Woods. Digital image processing, 2nd.\emph {SL: Prentice Hall}, 2002.

\bibitem{5}Caude Berge, Hypergraph: Combination of Finite Set, 1989.


\bibitem{6}Bretto, Alain, and Luc Gillibert. "Hypergraph-based image representation",\emph{ Graph-based Representations in Pattern Recognition}, Springer Berlin Heidelberg, pp. 1-11, 2005.



\bibitem{7}Stell, John G. "Relations on Hypergraphs", \emph{Relational and Algebraic Methods in Computer Science}, vol. 326, 2012.

\bibitem{8}Cousty, Jean, et al. "Morphological filtering on graphs", \emph{Computer Vision and Image Understanding},vol. 117, no. 4, pp. 370-385, 2013.
\bibitem{9}Bloch, Isabelle and Alain Bretto, "Mathematical Morphology on Hypergraphs: Preliminary Definitions and Results", \emph{Discrete Geometry for Computer Imagery}, pp. 429-440, Springer, 2011.
\bibitem{10}Sebastian, Bino, et al., "Mathematical Morphology on Hypergraphs Using Vertex-Hyperedge Correspondence," \emph{ISRN Discrete Mathematics}, 2014. 
\bibitem{11}Sebastian, V. Bino, et al., "Morphological filtering on hypergraphs," \emph{arXiv preprint arXiv:1402.4258}, 2014.
\bibitem{12}Frank  Y Shih, \emph{Image Processing and Mathematical Morphology: Fundamentals and Applications}, CRC press, 2010.

\bibitem{13}Jean Paul Serra, Image Analysis and Mathematical Morphology, 1982.
\bibitem{14}Dharmarajan, R., and K. Kannan. "Hypergraph-based edge detection in gray images by suppression of interior pixels."\emph{Global Journal of Science Frontier Research} vol. 12, no. 4, 2012

\bibitem{15}Stell, John G. "Formal concept analysis over graphs and hypergraphs."\emph{Graph Structures for Knowledge Representation and Reasoning} Springer International Publishing, pp. 165-179, 2014
\end{thebibliography}

\end{document}